\documentclass{article}

\usepackage{microtype}
\usepackage{graphicx}
\usepackage{subcaption}
\usepackage{booktabs} 

\usepackage{hyperref}

\usepackage[preprint]{icml2026}

\usepackage{amsmath}
\usepackage{amssymb}
\usepackage{mathtools}
\usepackage{amsthm}

\usepackage{algorithm}
\usepackage{algorithmic}

\usepackage[capitalize,noabbrev]{cleveref}

\theoremstyle{plain}

\theoremstyle{definition}

\theoremstyle{remark}

\icmltitlerunning{One Size does not Fit All: Heterogeneous Latent Space Alignment for Unsupervised Domain Adaptation}

\begin{document}

\twocolumn[
  \icmltitle{One Size does not Fit All: Heterogeneous Latent Space Alignment for Unsupervised Domain Adaptation}



  \icmlsetsymbol{equal}{*}

  \begin{icmlauthorlist}
    \icmlauthor{Evangelia Koskinioti}{Duke-ece}
    \icmlauthor{Yi Shen}{Duke-meng}
    \icmlauthor{Georgios Stamou}{Ntua}
    \icmlauthor{Michael M. Zavlanos}{Duke-meng}
  \end{icmlauthorlist}

  \icmlaffiliation{Duke-ece}{Department of Electrical and Computer Engineering, Duke University, NC, USA}
  \icmlaffiliation{Duke-meng}{Department of Mechanical Engineering, Duke University, NC, USA}
  \icmlaffiliation{Ntua}{Department of Electrical and Computer Engineering, National Technical University of Athens, Athens, Greece}

  \icmlcorrespondingauthor{Evangelia Koskinioti}{evangelia.koskinioti@duke.edu}

  \icmlkeywords{Domain Adaptation, Optimal Transport, Medical Imaging}

  \vskip 0.3in
]



\printAffiliationsAndNotice{}  

\begin{abstract}
\textit{Domain shift} remains a major obstacle to the reliable deployment of machine learning models in high-stakes environments such as healthcare. While \textit{Domain adaptation} aims to mitigate these effects, existing approaches suffer from limited expressiveness of latent representations and a reliance on handcrafted, static augmentations. In this work, we address these limitations by proposing a novel deep learning architecture for \textit{Unsupervised Domain Adaptation} (UDA), specifically optimized for medical image segmentation. Our framework, ADualVUOT, integrates a dual-encoder Variational Autoencoder (VAE) with Continuous Normalizing Flows (CNFs) to increase modeling flexibility and posterior expressiveness. To achieve domain alignment, we leverage \textit{Unbalanced Optimal Transport} (UOT) through the \textit{Gaussian-Gromov-Wasserstein} (GGW) distance, which handles structural and topological discrepancies between domains. Furthermore, we incorporate an adversarial augmentation scheme to synthesize worst-case compositions, thus enhancing model robustness. Extensive experiments on medical imaging benchmarks show significant gains over prior OT-based approaches.
\end{abstract}

\section{Introduction}

A fundamental challenge in machine learning is achieving reliable model generalization across diverse testing scenarios. This challenge is especially pronounced under \emph{domain shift}, where differences between the training and deployment data distributions lead to significant performance degradation \cite{domshift}.  Such performance failures can have serious consequences, particularly in healthcare applications such as medical imaging, where models must operate robustly across varying imaging devices, patient populations or acquisition protocols~\cite{domshift_medical}.
To address this issue, \textit{Domain Adaptation} (DA)~\cite{domadapt} aims to reduce the discrepancy between two given domains, the \emph{source/training} and the \emph{target/testing}, thus enabling effective knowledge transfer and generalization. \textit{Unsupervised Domain Adaptation} (UDA) is an even more challenging task, where target domain samples are available during training, but unlabeled.

A central mechanism in DA is the alignment of source and target representations in a latent space, which facilitates the transfer of knowledge learned from the source domain to the target \cite{project_Nguyen_2021_ICCV, project_GA}. Optimal Transport (OT) \cite{OT} has emerged as a popular framework for achieving domain alignment, in which deep neural networks project both domains into a common latent space while OT-based distances are incorporated into the training objective to minimize distributional discrepancies \cite{DeepJDOT, JUMBOT}. However, computing exact OT distances is prohibitively expensive, motivating approximations such as the Bures--Wasserstein distance \cite{Bures-W}, which models both domains as Gaussians and significantly reduces computational complexity. Despite these efficiency gains, such methods require source and target distributions to share the same latent dimensionality and rely on restrictive Gaussian assumptions. In practice, this can be limiting: source domains are often larger and more diverse, benefiting from higher-dimensional latent representations, whereas target domains—particularly in medical applications—are typically smaller, less complex, and frequently unlabeled \cite{alzheimers}. Enforcing alignment under these constraints can discard domain-specific information critical for downstream tasks, leading to degraded performance \cite{zhao2022fundamentallimitstradeoffsinvariant}.

In many real-world domain adaptation settings, especially in medical applications, performance is further hindered by the limited diversity of available training data. Medical datasets are commonly collected in localized clinical environments, where site-specific patient cohorts, imaging devices, and acquisition protocols yield narrowly distributed samples that fail to capture broader clinical and demographic variability \cite{domshift_medical2}. As a result, models trained on such data are highly susceptible to domain shifts at deployment. Data augmentation techniques have therefore been widely adopted to improve robustness and generalization \cite{sifa, maxstyle, hoffman2018cycada}. However, existing approaches typically rely on predefined transformations, prior knowledge of the target domain, or large-scale datasets, assumptions that are often unrealistic when the target domain is only partially observed, evolves over time, or when data availability is limited.

To address both the restrictive nature of shared latent representations and the limited generalization of handcrafted augmentation strategies, we propose \textbf{ADualVUOT} (Adversarial Dual Variational Unbalanced Optimal Transport), a novel framework for Unsupervised Domain Adaptation. Our approach decouples the latent representations of the source and target domains while performing geometry-aware alignment using the Gaussian Gromov--Wasserstein distance \cite{GGW}, enabling optimal transport between latent spaces of differing dimensionality. \textit{ADualVUOT} integrates a dual-encoder Variational Autoencoder with Continuous Normalizing Flows and Unbalanced Optimal Transport to learn expressive, domain-specific latent representations. In addition, we introduce an adversarial data augmentation scheme that dynamically generates challenging samples to improve robustness and facilitate the alignment process without requiring prior knowledge of the target domain or predefined augmentation policies. Extensive experiments on multiple medical image segmentation benchmarks demonstrate that \textit{ADualVUOT} consistently outperforms existing OT-based methods across a wide range of domain shifts.

\subsection{Statement of Contributions}
The main contributions of this work are summarized as follows:
\begin{itemize}
    \item We propose \textbf{ADualVUOT}, a novel Unsupervised Domain Adaptation framework that leverages Optimal Transport theory through the Gaussian Gromov--Wasserstein distance to perform geometry-aware alignment between heterogeneous latent spaces.
    \item We introduce a dual-encoder CNF-VAE architecture that projects source and target data into distinct latent spaces, enabling flexible modeling of domain-specific characteristics while avoiding restrictive shared-dimensionality assumptions. To the best of our knowledge, this is the first work to integrate a dual-encoder architecture with OT-based latent alignment for UDA.
    \item We incorporate an adversarial data augmentation strategy that dynamically generates challenging samples to improve robustness and facilitate alignment, without requiring prior knowledge of the target domain or predefined augmentation policies.
    \item We demonstrate the effectiveness of ADualVUOT on multiple medical image segmentation benchmarks, achieving consistent improvements over existing OT-based domain adaptation methods across diverse domain shifts.
\end{itemize}

\section{Related Work}

Unsupervised Domain Adaptation (UDA) focuses on transferring knowledge from a labeled source domain to an unlabeled target domain under distributional shift. We briefly review prior work most relevant to our approach, including data augmentation strategies and Optimal Transport (OT)-based latent space alignment.

\subsection{Data Augmentation for UDA}

Data augmentation has been widely used to improve robustness in UDA by perturbing source or target samples. Many approaches rely on generative models to translate source data into target-style samples, often using adversarial learning \cite{sankaranarayanan2018generateadaptaligningdomains, hoffman2018cycada, russo2017sourcetargetbacksymmetric}. Related strategies have also been explored in domain generalization, where target data is unavailable during training \cite{li2021progressivedomainexpansionnetwork, maxstyle}. In medical imaging, SIFA \cite{sifa} combines image- and feature-level alignment using cycle-consistent translation. Despite their effectiveness, such methods often depend on adversarial training and sufficient data diversity, which can be challenging in medical settings. In contrast, our approach employs a minimax optimization strategy that does not rely on
GAN-based image translation or domain discriminators. Adversarial training is instead
confined to feature-level perturbations parameterized by simple Gaussian noise, which
requires fewer data assumptions while still directly challenging the segmentation objective.
This design enables robust adaptation without introducing additional generative
networks or complex adversarial pipelines.

\subsection{Latent Space Alignment via Optimal Transport}

A large body of UDA methods aligns source and target distributions in a shared latent space using discrepancy measures. Optimal Transport is widely adopted for this purpose, owing to its strong theoretical foundations. JDOT \cite{JDOT} and its deep extension DeepJDOT \cite{DeepJDOT} align source and target distributions using the Wasserstein distance, while JUMBOT \cite{JUMBOT} incorporates Unbalanced Optimal Transport to address class imbalance and support mismatch. More structured latent spaces have been explored using Variational Autoencoders, as in OLVA \cite{OLVA}, and improved through the Bures--Wasserstein distance in PUFT \cite{PUFT}. Our work builds upon these approaches by enabling alignment across heterogeneous latent spaces using the Gromov--Wasserstein distance.
\section{Method}
In this section, we introduce our proposed architecture, \textbf{ADualVUOT} (Adversarial Dual Variational Unbalanced Optimal Transport), for unsupervised domain adaptation in medical image segmentation. Our framework integrates four key components: (1) a dual-encoder VAE that projects source and target samples into two distinct, Gaussian-structured latent spaces (Section \ref{subsec:dualvae}); (2) two separate CNF architectures that transform these latent variables to model more expressive posteriors (Section \ref{subsec:cnfvae}); (3) a UOT-based alignment component designed to minimize domain discrepancy within the latent space (Section \ref{subsec:HOT}); and (4) MaxStyle \cite{maxstyle}, an adversarial feature augmentation method that perturbs images from both domains to promote the learning of domain-invariant features, adjusted for domain adaptation (Section \ref{subsec:max_style_uda}). 

\begin{figure*}[t]
  \centering
  \includegraphics[width=\textwidth]{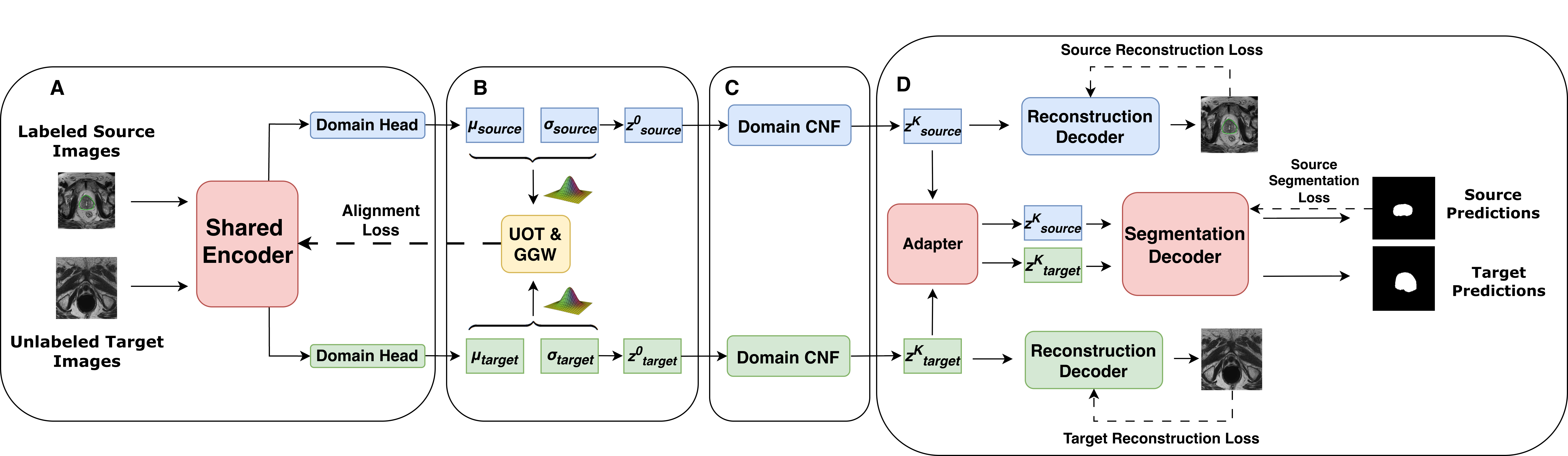}
  \caption{
    Illustration of the DualVUOT proposed architecture. \textbf{Part A:} Source and target images pass through the shared encoder with domain-specific heads, producing Gaussian posterior parameters. \textbf{Part B:} The UOT-GGW module uses these to produce the domain alignment loss. \textbf{Part C:} The latent representations pass through domain-specific CNFs that map them to a more expressive posterior. \textbf{Part D:} Transformed embeddings are decoded for reconstruction and segmentation; source labels supervise segmentation, while target pseudo-labels support subsequent alignment.}
  \label{fig:dualvuot}
\end{figure*}

\subsection{Notation}

Following the standard UDA setting, we denote the labeled source domain dataset as $\mathcal D^s=\{(x_i^s,y_i^s)\}_{i=1}^n$ and the unlabeled target domain dataset as $\mathcal D^t=\{x_j^t\}_{j=1}^m$.

\subsection{Dual-Encoder Variational Autoencoder with Continuous Normalizing Flows}
\label{subsec:3.2}

\subsubsection{Dual-Encoder VAE}
\label{subsec:dualvae}
Let $d\in\{s,t\}$ denote the domain. A shared encoder, \emph{Enc}, for both domains maps the input images to a shared feature representation $h = \text{Enc}(x^d) \in \mathbb{R}^H$. Then, the encoder splits into domain-specific, fully-connected heads, \emph{FC}, that predict the parameters of two distinct Gaussian posteriors:
\begin{align*}
\mu_d(x^d) &= \text{FC}_{\mu^d}(h), \quad
\sigma_d(x^d) = \text{FC}_{\sigma^d}(h),
\end{align*}
where $\mu_s(x^s), \sigma_s(x^s) \in \mathbb{R}^K, \mu_t(x^t), \sigma_t(x^t) \in \mathbb{R}^P$, with latent dimensions $K$ and $P$ respectively. The shared encoder encourages representation similarity prior to alignment, while the separate heads enable the encoding of domain-specific information.
Then, using the reparameterization trick in \cite{Kingma_2019}, the latent space representation can be sampled as:
\begin{equation*}
    z_0^d = \mu_d(x^d) + \sigma_d(x^d) \odot \epsilon, \quad \epsilon \sim \mathcal{N}(0, I),
\end{equation*}
with the resulting approximate posterior:
\begin{equation*}
q_{_d}(z^d\mid x^d)=\mathcal{N}\!\big(z^d;\mu_d(x^d),\mathrm{diag}(\sigma_d^2(x^d))\big)
\end{equation*}
Although valid, the Gaussian assumption imposes a restrictive inductive bias on the latent space, so we treat $z_0^d$ as a \emph{base latent variable}, which is subsequently transformed into a more expressive representation through a Continuous Normalizing Flow (CNF).

\subsubsection{Continuous Normalizing Flow-Based VAE}
\label{subsec:cnfvae}
Formally, the CNF defines a continuous-time transformation of the latent variable, $z^d(t)$, initialized at $z^d(0) = z_0^d$, which evolves until time $T$ and yields a more expressive latent representation $z_T^d = z^d(T)$. According to~\cite{cnf}, the density of the distribution of the transformed variable $z^d_T$ can be calculated as:
\begin{equation}
    \label{eq:cnf_loglikelihood}
    \log p(z^d_T) = \log p(z^d_0) - \int_{0}^{T} \text{Tr}\Big(\frac{\partial f}{\partial z^d(t)}\Big) dt
\end{equation}
where $f$ is a neural network that defines a time-dependent vector field over the latent
space. In this work, we employ the family of planar normalizing flows \cite{varnf}, which define a series of non-linear transformations of the form:
\begin{equation*}
    z^d(t+1) = f(z^d(t)) = z^d(t) + u \cdot \text{act}(w^T\cdot z^d(t)+b), 
\end{equation*}
where ``act'' is a non-linear activation function and $u, w$ and $b$ are learnable parameters. Further details on this section can be found in the Appendix, Section \ref{appendix:cnfvae}.

\subsubsection{Domain-Specific Training Objectives}
\label{subseq:losses}
Finally, the CNF-transformed latent variables $z_T^s,z_T^t$ are decoded for domain-specific image reconstruction, ensuring the preservation of the input's vital information, while $z_T^s$ is also used for supervised segmentation. We derive the domain-wise training objectives used to jointly optimize the VAE, CNF, reconstruction and segmentation components.

Let $p^d_\omega(x^d | z^d_T)$ denote the domain-specific reconstruction decoders. Assuming a standard Gaussian prior $p(z) \sim \mathcal{N}(0,I)$, we optimize the model using the Evidence Lower Bound (ELBO) \cite{Kingma_2019} for each domain $d\in\{s,t\}$. 
\begin{equation*}
\begin{aligned}
{\mathrm{ELBO}}^{(d)}(x^d)
&=
\mathbb{E}_{q_d(z_T^d \mid x^d)}
\big[
\log p^d_\omega(x^d \mid z_T^d)
\big]
\\
&\quad
-
\mathrm{KL}
\big(
q_d(z_T^d \mid x^d)
\;\|\;
p(z)
\big)
\end{aligned}
\end{equation*}
Replacing the log-likelihood of the transformed variable from equation \eqref{eq:cnf_loglikelihood} we get the ELBO derivation for a VAE-CNF:
\begin{equation*}
\begin{aligned}
{\mathrm{ELBO}}^{(d)}(x^d)
=
\mathbb{E}_{q_{0,d}(z_0^d \mid x^d)}
\Big[
\log p^d_\omega(x^d \mid z_T^d)
+
\log p(z_T^d)
\\
\quad
-
\log q_{d}(z_0^d \mid x^d)
+
\int_0^T
\mathrm{Tr}
\!\left(
\frac{\partial f}{\partial z^d(t)}
\right)
dt
\Big]
\end{aligned}
\end{equation*}
Equivalently, we use:
  $  \mathcal{L}_{\mathrm{ELBO}}^{(d)}(x^d) = - {\mathrm{ELBO}}^{(d)}(x^d)$.

Since the target images are unlabeled, segmentation supervision is applied exclusively for source images through a segmentation loss $\mathcal{L}_\text{seg}$. We denote the predicted source labels as $\hat{y}^s$, leading to the following domain-specific objectives:
\begin{equation}
    \begin{aligned}
        \label{eq:domain_losses}
        \mathcal{L}^{(s)}(x^s, y^s) &= \mathcal{L}^{(s)}_{\mathrm{ELBO}}(x^s) + \lambda_{\mathrm{seg}}\mathcal{L}_{\mathrm{seg}}(y^s,\hat{y}^s), \\
\mathcal{L}^{(t)}(x^t)      &= \mathcal{L}_{\mathrm{ELBO}}^{(t)}(x^t)
    \end{aligned}
\end{equation}
\subsection{Hierarchical Unbalanced Optimal Transport Loss with the Gaussian Gromov-Wasserstein Distance}
\label{subsec:HOT}

Our Dual-Encoder VAE-CNF architecture learns expressive and meaningful representations for both domains, but expressiveness alone does not guarantee cross-domain alignment. To achieve alignment, we introduce an Optimal Transport-based alignment loss computed in the Gaussian latent spaces. In the absence of a shared latent space, we employ a \emph{hierarchical} OT formulation based on the Gaussian Gromov-Wasserstein (GGW) distance \cite{GGW}. The hierarchy arises from two levels of alignment: (i) a mini-batch level alignment that establishes correspondences between similar source and target latent representations via UOT and (ii) an embedding-level alignment via GGW, where latent embeddings are modeled and compared as structured distributions.
\subsubsection{Mini-Batch Level Alignment using UOT}
\label{subsec:miniuot}
Let $\mu$ and $\nu$ denote empirical probability measures defined over
source and target \emph{mini-batches} of size $B$, $\{z_{0,i}^s\}_{i=1}^B$ and $\{z_{0,j}^t\}_{j=1}^B$ respectively:
\begin{align}
  \mu = \sum_{i=1}^B a_i \, \delta_{z_{0,i}^s},
\qquad
\nu = \sum_{j=1}^B b_j \, \delta_{z_{0,j}^t}  
\end{align}
with uniform weights $a_i = b_i = \frac{1}{B}$ and $\delta_{z^d_0}$ denoting a Dirac measure centered at $z_0^d$. Given these distributions, we compute the Unbalanced Optimal Transport plan $\gamma^* \in \mathbb{R}^{B \times B}$, which defines a soft correspondence between the source and target samples: 
\begin{equation}
    \label{eq:uot_plan}
    \gamma^* = \arg \min_{\gamma \geq 0} \; \langle \gamma, C \rangle+ \rho_s D(\gamma \mathbf{1}||a) + \rho_t D(\gamma^\top \mathbf{1}||b)
\end{equation}
where $C \in \mathbb{R}^{B \times B}$ is the transportation cost matrix, $D(\cdot\|\cdot)$ denotes a divergence measure (e.g., KL divergence), and $\rho_s,\rho_t$ control the degree of mass relaxation. In our framework, the transport plan is determined by a pairwise embedding cost $C_{ij}$, which we detail in the following paragraph.

\subsubsection{Embedding Level Alignment using GGW}
At the embedding level, each latent representation $z_0^d$ is modeled as a Gaussian distribution. The GGW distance allows us to perform transport between distributions defined in different metric spaces (latent dimensions), quantifying their similarity by comparing their internal geometric structure rather than relying on pointwise distances. Given a single source $z_0^s \sim\mathcal{N}(\mu_s, \Sigma_s) \in \mathbb{R}^K$ and a target $z_0^t \sim \mathcal{N}(\mu_t, \Sigma_t) \in \mathbb{R}^P$ embedding, the GGW distance is defined as:
\begin{equation*}
\begin{aligned}
&\mathrm{GGW}_2^2(z_0^s, z_0^t)
=
4\Big(\operatorname{Tr}(D_s) - \operatorname{Tr}(D_t)\Big)^2 \\
& \quad 
+ 8\big\|D_s^{(P)} - D_t\big\|_F^2 
+ 8\Big(\|D_s\|_F^2 - \|D_s^{(P)}\|_F^2\Big)
\end{aligned}
\end{equation*}
where $D_s$ and $D_t$ are diagonal matrices containing the eigenvalues of $\Sigma_s$ and $\Sigma_t$, sorted in decreasing order and $\|\cdot\|_F$ denotes the Frobenius norm. Assuming $K>P$, $D_s^{(P)}$ contains only the leading $P$ eigenvalues of $D_s$. Therefore, the distance between two embeddings $i$ and $j$ is defined as:
\begin{equation*}
\label{eq:embed_cost}
C^{\mathrm{embed}}_{ij}
=
\mathrm{GGW}_2^2\!\left(
z_{0,i}^s,
z_{0,j}^t
\right)
\end{equation*}
However, effective domain adaptation also requires semantic consistency across domains to prevent geometrically similar but semantically mismatched embeddings from being coupled during transport. Since target images are unlabeled, semantic similarity is defined based on the source labels and target pseudo-labels predicted by the segmentation decoder, denoted as $\hat{y}^t$. We define the semantic component of the transport cost using the Dice Similarity Coefficient (DSC):
\begin{equation*}
    C^{\mathrm{dice}}_{ij} = 1-\text{DSC}(y^s_i,\hat{y}^t_j)
\end{equation*}
Finally, the transport cost between a source sample $i$ and a target sample $j$ is defined
as a weighted combination of these two components:
\begin{equation}
\label{eq:costm}
C_{ij}
=
\lambda_{\text{embed}} \, C^{\mathrm{embed}}_{ij}
+
\lambda_{\text{dice}}\, C^{\mathrm{dice}}_{ij}
\end{equation}
\subsubsection{Unbalanced Optimal Transport Transfer Loss}
\label{subsec:uot_loss}
Given the UOT plan $\gamma^*$ (eq.~\eqref{eq:uot_plan}) based on the cost matrix $C$ (eq.~\eqref{eq:costm}), the UOT transfer loss is:
\begin{equation}
    \label{eq:transferloss_final}
    \mathcal{L}_{\text{UOT}} = \sum_{i=1}^B \sum_{j=1}^B \gamma^*_{ij} \cdot C_{ij}
\end{equation}
Minimizing this objective encourages the model to produce embeddings that are geometrically close in the latent space, and especially for samples that are semantically similar. The optimal coupling $\gamma^*$ is computed for a pair of mini-batches by solving eq.~\eqref{eq:uot_plan} for a given $C$ using iterative Sinkhorn updates~\cite{cuturi2013sinkhorn}, and then used to evaluate the transfer loss. 
\subsection{DualVUOT for Unsupervised Domain Adaptation}
Our DualVUOT framework for image segmentation in UDA, illustrated in Figure~\ref{fig:dualvuot}, is trained using the following objective, combining equations~\eqref{eq:domain_losses} and~\eqref{eq:transferloss_final}:
\begin{equation}
\label{eq:dualvuot_loss}
    \begin{aligned}
        \mathcal{L}_{\text{total}}(x^s,y^s,x^t)
&=
\lambda_s\,\mathbb{E}_{(x^s,y^s)\sim\mathcal{D}_s}
\big[\mathcal{L}^{(s)}(x^s,y^s)\big] \\
&
+
\lambda_t\,\mathbb{E}_{x^t\sim\mathcal{D}_t}
\big[\mathcal{L}^{(t)}(x^t)\big] 
+ \lambda_{\mathrm{UOT}}\,\mathcal{L}_{\mathrm{UOT}}
    \end{aligned}
\end{equation}
\begin{figure*}[t]
  \centering
  \includegraphics[width=\textwidth]{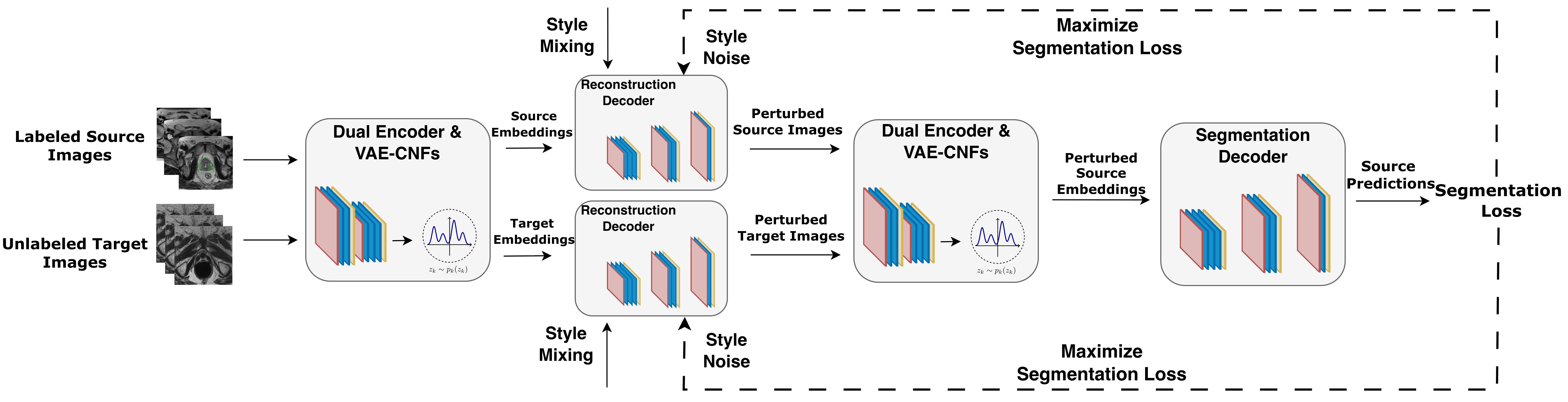}
  \caption{
    Illustration of our final proposed architecture, ADualVUOT. For simplicity, the Dual-Encoder and CNF modules are visualized as a black-box that produces the latent space representations, since they are already illustrated in detail in Figure \ref{fig:dualvuot}. The training pipeline for this architecture is also explained in Algorithm \ref{alg:ADualVUOT}.
  }
  \label{fig:ours}
\end{figure*}
\subsection{Adversarial Dual VUOT (ADualVUOT) for Unsupervised Domain Adaptation}
\label{subsec:max_style_uda}
To further mitigate the impact of domain shift, we utilize MaxStyle \cite{maxstyle}, an adversarial feature augmentation method originally proposed for domain generalization, and adapt it to the UDA setting, to get our final proposed model, \textbf{ADualVUOT}. Our goal is to create meaningful and challenging augmentations of both source and target images that outperform standard augmentation strategies.
First, we mix the styles of the source images with the target images and vice versa, \emph{within their respective reconstruction decoders}, following the formulation of \cite{maxstyle}. More specifically, let a source and a target image $x_i^s, x_j^t$ and their $c$-dimensional feature maps extracted by an intermediate layer of their respective reconstruction decoders $f_{i}^s, f_{j}^t \in \mathbb{R}^{c \times h \times w}$. Also, let the channel-wise means and standard deviations of the feature maps be defined as $\mu(f^s_i), \sigma(f^s_i)$ and $\mu(f^t_j), \sigma(f^t_j)$, respectively. To mix the styles of the two images, we transform each feature map with a linear combination of the other domain's style statistics:
\begin{align*}
    \text{Mixed}(f_{i}^s) &= \gamma_{mix}^s \odot f_{i}^s + \beta_{mix}^s, \quad \text{where} \\
    \gamma_{mix}^s &= \lambda_{mix}\, \sigma(f_{i}^s) + (1 - \lambda_{mix})\, \sigma(f_{j}^t) \\
    \beta_{mix}^s  &= \lambda_{mix}\, \mu(f_{i}^s)    + (1 - \lambda_{mix})\, \mu(f_{j}^t)
\end{align*}
\begin{align*}
    \text{Mixed}(f_{j}^t) &= \gamma_{mix}^t \odot f_{j}^t + \beta_{mix}^t, \quad \text{where} \\
    \gamma_{mix}^t &= \lambda_{mix}\, \sigma(f_{j}^t) + (1 - \lambda_{mix})\, \sigma(f_{i}^s) \\
    \beta_{mix}^t  &= \lambda_{mix}\, \mu(f_{j}^t)    + (1 - \lambda_{mix})\, \mu(f_{i}^s)
\end{align*}
where $\lambda_{mix} \in [0,1]$ controls the degree of interpolation for the 
source and target branches respectively.

Then, to further expand the style space, we follow \cite{maxstyle} and introduce additive style noise, which produces the augmented feature maps:
\begin{align*}
    \tilde{f}_{i}^{s} &= \left(\gamma_{mix}^s + \Sigma_\gamma \cdot \epsilon_\gamma\right)
                          \odot f_{i}^s
                        + \left(\beta_{mix}^s  + \Sigma_\beta  \cdot \epsilon_\beta\right) \\
    \tilde{f}_{j}^{t} &= \left(\gamma_{mix}^t + \Sigma_\gamma \cdot \epsilon_\gamma\right)
                          \odot f_{j}^t
                        + \left(\beta_{mix}^t  + \Sigma_\beta  \cdot \epsilon_\beta\right)
\end{align*}
where  $\epsilon_\gamma, \epsilon_\beta \sim \mathcal{N}(0, 1)$ and $\Sigma_\gamma = \mathrm{Var}\left({\sigma(f_{k}^s)}_{k=1}^{B}\right)$, $\Sigma_\beta  = \mathrm{Var}\left({\mu(f_{k}^s)}_{k=1}^{B}\right)$. After applying the style mixing and additive noise, each reconstruction decoder continues from its respective perturbed feature map and produces a perturbed version of the input image, $\tilde{x}_i^s$ and $\tilde{x}_j^t$. The perturbed source image is also used to generate the new prediction $\hat{\tilde{y}}_i^s$, for the calculation of the source segmentation loss $\mathcal{L}_{seg}(y_i^s,\hat{\tilde{y}}_i^s)$. 

In order to generate meaningfully challenging augmentations, we optimize the noise parameters 
$\epsilon_\gamma, \epsilon_\beta$ and the source mixing coefficient $\lambda_{mix}$ via 
gradient ascent on the \textbf{source} segmentation loss, keeping the network weights frozen 
during this inner loop:
\begin{align*}
    \label{eq:adv_opt_source}
    \epsilon_\gamma   &\leftarrow \epsilon_\gamma 
                           + \alpha\, \nabla_{\epsilon_\gamma}\, 
                             \mathcal{L}_{seg}(y_i, \hat{\tilde{y}}_i) \nonumber \\
    \epsilon_\beta   &\leftarrow \epsilon_\beta  
                           + \alpha\, \nabla_{\epsilon_\beta}\,  
                             \mathcal{L}_{seg}(y_i, \hat{\tilde{y}}_i) \nonumber \\
    \lambda_{mix}     &\leftarrow \mathrm{Clip}_{[0,1]}\!\left(
                             \lambda_{mix} 
                           + \alpha\, \nabla_{\lambda_{mix}}\, 
                             \mathcal{L}_{seg}(y_i, \hat{\tilde{y}}_i)\right)
\end{align*}
where  $\alpha$ is the step size. This process is repeated for $K$ steps, after which the optimized parameters $\epsilon_\gamma^{*},\, \epsilon_\beta^{*},\, \lambda_{mix}^{*}$  define the worst-case style compositions. Each reconstruction decoder then 
produces the final style-augmented images $x_i^{s*}$ and $x_j^{t*}$ using these optimized 
parameters, which are passed through the full network a second time to obtain the final augmented segmentation predictions $\hat{y}_i^{s*}$.
Finally, the ADualVUOT architecture is trained by minimizing the clean objective together with
the corresponding worst-case style-augmented objective, following equations~\eqref{eq:domain_losses},\eqref{eq:dualvuot_loss}:
\begin{equation*}
\label{eq:adualvuot_loss}
\begin{aligned}
\min_{\theta}\;
\mathcal{L}
&=
\mathcal{L}_{\mathrm{total}}(x^s,y^s,x^t)+
\lambda_{\mathrm{adv}}
\mathcal{L}_{\mathrm{total}}(x^{s*},y^s,x^{t*})
\end{aligned}
\end{equation*}
\begin{equation*}
\text{s.t.} \quad
\epsilon_\gamma^{*},\, \epsilon_\beta^{*},\, \lambda_{mix}^{*}
=
\operatorname*{arg\,max}_{\epsilon_\gamma,\, \epsilon_\beta,\, \lambda_{mix}}
\mathcal{L}_{\mathrm{seg}}(y^s,\hat{\tilde{y}}^s)
\end{equation*}
The architecture is illustrated in Figure~\ref{fig:ours} and the training process is explained 
step by step in Algorithm~\ref{alg:ADualVUOT}.
\begin{algorithm}[t]
\caption{ADualVUOT for UDA}
\label{alg:ADualVUOT}
\begin{algorithmic}[1]
\STATE {\bfseries Input:} source dataset $\mathcal{D}^s=\{(x_i^s,y_i^s)\}_{i=1}^n$, 
                          target dataset $\mathcal{D}^t=\{x_j^t\}_{j=1}^m$
\STATE {\bfseries Model:} Dual-Encoder VAE-CNF, reconstruction decoders $p_{\omega_s},p_{\omega_t}$, segmentation decoder $p_{\omega_{\mathrm{seg}}}$

\FOR{each source-target mini-batch $(x^s,y^s)$ and $x^t$}

  \STATE Compute clean forward pass and clean loss $\mathcal{L}_{\mathrm{total}}(x^s,y^s,x^t)$

  \STATE Extract intermediate reconstruction-decoder features $f^s$ and $f^t$ at layer $\ell$

  \STATE Initialize $\epsilon_\gamma,\epsilon_\beta \sim \mathcal{N}(0,1)$ and $\lambda_{mix}\sim \mathrm{Uniform}(0,1)$

  \STATE {\bfseries Inner Optimization Loop ($K$ steps), network weights frozen:}
  \FOR{$k = 1 \text{ to } K$}

    \STATE Compute mixed style statistics from $f^s$ and $f^t$
    \STATE Apply style mixing and noise to obtain $\tilde{f}^s,\tilde{f}^t$
    \STATE Decode $\tilde{f}^s,\tilde{f}^t$ to obtain $\tilde{x}^s,\tilde{x}^t$
    \STATE Run the full network on $\tilde{x}^s$ and obtain $\hat{\tilde{y}}^s$

    \STATE Update adversarial style parameters by gradient ascent:
    \STATE $\epsilon_\gamma \leftarrow \epsilon_\gamma 
            + \alpha\nabla_{\epsilon_\gamma}\mathcal{L}_{\mathrm{seg}}(y^s,\hat{\tilde{y}}^s)$
    \STATE $\epsilon_\beta \leftarrow \epsilon_\beta 
            + \alpha\nabla_{\epsilon_\beta}\mathcal{L}_{\mathrm{seg}}(y^s,\hat{\tilde{y}}^s)$
    \STATE $\lambda_{mix} \leftarrow \mathrm{Clip}_{[0,1]}\!\left(
            \lambda_{mix}
            + \alpha\nabla_{\lambda_{mix}}\mathcal{L}_{\mathrm{seg}}(y^s,\hat{\tilde{y}}^s)
            \right)$

  \ENDFOR

  \STATE Generate worst-case augmented inputs $x^{s*},x^{t*}$ using $\epsilon_\gamma^*,\epsilon_\beta^*,\lambda_{mix}^*$

  \STATE Compute augmented loss $\mathcal{L}_{\mathrm{total}}(x^{s*},y^s,x^{t*})$

  \STATE Update model parameters by minimizing:
  \STATE $\mathcal{L}
  =
  \mathcal{L}_{\mathrm{total}}(x^s,y^s,x^t)
  +
  \lambda_{\mathrm{adv}}\mathcal{L}_{\mathrm{total}}(x^{s*},y^s,x^{t*})$

\ENDFOR
\end{algorithmic}
\end{algorithm}
\section{Experiments and Results}
\begin{table*}[t]
  \begin{center}
    \begin{small}
      \begin{sc}
        \begin{tabular}{lccccccc}
          \toprule
          Method & A & B & C & D & E & F & OOD \\
          \midrule
          ADualVUOT (ours) & \textbf{0.923} & \textbf{0.878} & \textbf{0.828} & \textbf{0.869} & \textbf{0.781} & \textbf{0.818} & \textbf{0.849} \\
          DualVUOT (ours)& 0.917 & 0.852 & 0.787 & 0.836 & 0.745 & 0.784 & 0.820 \\
          VUGW (ours)    & 0.911 & 0.803 & 0.759 & 0.794 & 0.689 & 0.755 & 0.785 \\
          PUFT  \cite{PUFT} & 0.920 & 0.810 & 0.741 & 0.780 & 0.710 & 0.763 & 0.787 \\
          JUMBOT \cite{JUMBOT} & 0.893 & 0.728 & 0.728 & 0.796 & 0.611 & 0.718 & 0.746 \\
          NoAdaptation   & 0.693 & 0.388 & 0.352 & 0.401 & 0.135 & 0.367 & 0.389 \\
          \bottomrule
        \end{tabular}
      \end{sc}
    \end{small}
    \vskip 0.1in
      \caption{Dice scores across different sites and out-of-distribution (OOD) evaluation for the Multi-Site Prostate MRI Dataset. Columns A--F represent the six distinct acquisition sites used for evaluation. OOD denotes the average Dice score across all sites, serving as an out-of-distribution performance metric.}
  \label{tab:ood_results}
  \end{center}
  \vskip -0.1in
\end{table*}
In this section, we detail our implementation and present experimental results for our method across the following public medical imaging benchmarks, specifically designed for unsupervised domain adaptation in image segmentation.
\subsection{Datasets}
\subsubsection{Multi-Site Prostate MRI Segmentation Dataset \cite{prostate}}
The Multi-Site Prostate MRI dataset consists of T2-weighted MRI scans with
corresponding segmentation masks acquired across seven sites (A--G) from different
institutions and challenges \cite{bloch2015nci,lemaitre2015computer,litjens2014evaluation}.
Following standard practice \cite{liu2021feddg}, site G is used as the source domain,
while sites A--F serve as target domains. Variations in scanners and acquisition
protocols introduce substantial domain shifts in image appearance, contrast, and noise.
\subsubsection{Multi-Modality Whole Heart Segmentation Dataset \cite{mmwhs}}
The MM-WHS dataset provides annotated cardiac volumes for MRI and CT modalities,
enabling evaluation of cross-modality adaptation. It contains 40 training volumes
(20 per modality) with labels for four cardiac structures: ascending aorta (AA), left
atrium (LA), left ventricle (LV), and myocardium (Myo). Adaptation between MRI and CT
is challenging due to modality-specific feature distributions and non-aligned data
manifolds \cite{zhuang2016multi,zhuang2019evaluationalgorithmsmultimodalityheart}.
\subsection{Implementation Details}
The shared encoder trunk consists of four residual downsampling blocks with [64, 128, 256, 512] channels, followed by two separate fully connected layers that lead to the domain-specific latent spaces. All the decoders mirror the encoders' structure, consisting of four residual upsampling blocks with [512, 256, 128, 64] channels, with the segmentation decoder ending with a softmax activation, while the reconstruction decoders end with a sigmoid activation. Both planar CNFs consist of 3 hidden layers.

The latent space dimensions were $K=128$ for the source domain and $P=64$ for the target domain. The source domain contains richer supervisory information and serves as the primary source of transferable knowledge, which is why we allocate greater representational capacity to it, while encouraging a more compact target representation. This design is consistent with our objective of learning expressive source embeddings while preserving only the target information necessary for cross-domain alignment.

All models were implemented with PyTorch for 200 epochs using an Adam optimizer with a learning rate and weight decay of $10^{-4}$. For our experiments, we used a batch size of 20. Across all experiments, we set the weights of the costs of equation~\eqref{eq:costm} as $\lambda_{\mathrm{embed}}=0.01$ and $\lambda_{\mathrm{dice}}=0.5$.

We evaluate all methods using the Dice Similarity Coefficient (DSC), and compare our approach against several OT-based baselines from the literature, including JUMBOT \cite{JUMBOT} and PUFT \cite{PUFT}. In addition, we conduct ablation studies using two variants of our model: \textit{VUGW}, which replaces the proposed UOT-GGW alignment with an Unbalanced Gromov-Wasserstein transport plan (Appendix~\ref{appendix:ugw}), and \textit{DualVUOT}, which removes the adversarial augmentation component to isolate its contribution. Finally, we include a source-only baseline (\textit{NoAdaptation}) to quantify the degree of domain shift and the necessity of domain adaptation.
\subsection{Results}
\subsubsection{Multi-Site Prostate MRI Segmentation Dataset}
The segmentation results on the Multi-Site Prostate MRI dataset are reported in Table~\ref{tab:ood_results}. Using site G as the source domain, each column corresponds to the model results on each target site. Compared to all other methods, our proposed \textit{ADualVUOT} consistently achieves the highest Dice scores across all target domains. The NoAdaptation baseline highlights substantial domain shift across sites, with site A being the least challenging and site E the most difficult. While improvements on site A are marginal, due to already high performance ($>90\%$ Dice), our method yields substantial gains on more challenging sites, particularly D and E, with improvements of up to $8\%$ and $7\%$, respectively, over PUFT, the strongest prior baseline.

We further analyze the contribution of individual components through a short ablation
study. Comparing ADualVUOT to DualVUOT demonstrates the benefit of the adversarial augmentation scheme, which leads to consistently improved performance. This effect is particularly pronounced for this dataset, where variations in acquisition sites and scanners introduce complex appearance shifts that are difficult to capture using static augmentation strategies alone. Moreover, comparing VUGW to DualVUOT indicates that the Gromov–Wasserstein distance is more effective when incorporated hierarchically within the transport cost.

\subsubsection{Multi-Modality Whole Heart Segmentation}
This dataset encompasses two adaptation tasks: MRI-to-CT and CT-to-MRI. As observed in our experiments, the adaptation from CT to MRI is more challenging than the inverse, likely because the MRI modality captures a higher degree of information and variability. We apply all previously mentioned methods to both tasks and report the results in Tables \ref{tab:mri_to_ct} and \ref{tab:ct_to_mri}.
\begin{table}[t]
  \centering
  \setlength{\tabcolsep}{4pt}
  \begin{small}
    \begin{tabular}{lccccc}
      \toprule
      Method & AA & LA & LV & Myo & OOD \\
      \midrule
      ADualVUOT (ours) & \textbf{0.931} & \textbf{0.927} & \textbf{0.911} & \textbf{0.782} & \textbf{0.888} \\
      DualVUOT (ours)& 0.909 & 0.904 & 0.903 & 0.765 & 0.870 \\
      VUGW  (ours)   & 0.873 & 0.857 & 0.888 & 0.754 & 0.843 \\
      PUFT           & 0.881 & 0.885 & 0.875 & 0.741 & 0.846 \\
      JUMBOT         & 0.626 & 0.617 & 0.607 & 0.592 & 0.611 \\
      NoAdaptation   & 0.190 & 0.218 & 0.290 & 0.136 & 0.209 \\
      \bottomrule
    \end{tabular}
  \end{small}
  \vskip 0.1in
  \caption{Dice Scores for MM-WHS (MRI $\rightarrow$ CT adaptation)}
  \label{tab:mri_to_ct}
\end{table}
For both tasks, our proposed ADualVUOT architecture outperforms all other approaches, including our own baseline. The improvement is particularly significant for the CT-to-MRI adaptation task, where we observe a 5\% overall increase compared to PUFT, the strongest baseline from previous literature. Notably, for the AA and Myo classes, our method achieves a 7\% improvement over PUFT, bringing the Dice scores for all classes above 70\%. During our ablation study we also observe that DualVUOT also improves over PUFT with an average 3\% increase, a margin that is further extended by the addition of the adversarial augmentation scheme. Also, comparing VUGW to DualVUOT demonstrates the benefit of the dual-encoder design combined with GGW-based alignment in the embedding space, since DualVUOT consistently has superior performance.
For the MRI-to-CT task, we also report an improvement over all other methods, outperforming PUFT by up to 4\%. Even for the easier classes, such as AA, in which PUFT already achieved a performance of 88\%, the Dice score increased by 5\%, while for all other classes, the improvement remained consistently over 4\%. 
In this experiment, DualVUOT yielded the most significant impact, achieving substantial performance gains, while the adversarial augmentation scheme further closes the gap. By providing the more diverse source domain with a higher-dimensional latent space and allowing the more limited target domain to be further compressed, the model is able to learn the downstream task more effectively. 
\begin{table}[t]
  \centering
  \setlength{\tabcolsep}{4pt}
  \begin{small}
    \begin{tabular}{p{2.4cm}ccccc}
      \toprule
      Method & AA & LA & LV & Myo & OOD \\
      \midrule
      ADualVUOT (ours) & \textbf{0.767} & \textbf{0.821} & \textbf{0.857} & \textbf{0.704} & \textbf{0.787} \\
      DualVUOT (ours)& 0.744 & 0.791 & 0.836 & 0.672 & 0.761 \\
      VUGW (ours)    & 0.723 & 0.769 & 0.812 & 0.651 & 0.739 \\
      PUFT           & 0.693 & 0.774 & 0.830 & 0.636 & 0.733 \\
      JUMBOT         & 0.602 & 0.514 & 0.741 & 0.390 & 0.562 \\
      NoAdaptation   & 0.020 & 0.202 & 0.075 & 0.127 & 0.106 \\
      \bottomrule
    \end{tabular}
  \end{small}
  \vskip 0.1in
  \caption{Dice Scores for MM-WHS (CT $\rightarrow$ MRI adaptation)}
  \label{tab:ct_to_mri}
\end{table}
\section{Conclusions}
In this work, we proposed ADualVUOT, a novel deep learning architecture for semantic segmentation under unsupervised domain adaptation. We extended the existing VAE-CNF architecture by incorporating a dual-encoder design that allowed the source and target domains to be projected into latent spaces of different dimensionalities, thus enabling greater modeling flexibility and avoiding excessive compression or sparsity in the latent space representations. We also introduced the GGW distance as a powerful alignment metric that mitigates domain discrepancy by preserving the geometric and structural properties of each latent space, thus allowing for more expressive alignment. Lastly, we adapted the \textit{MaxStyle} adversarial augmentation scheme for the domain adaptation setting, which enhanced model robustness and bypassed the need for static, handcrafted augmentations. We presented experimental results that showed that our method achieved state-of-the-art results on popular medical imaging benchmarks, with each component contributing substantially to performance improvement.  

\bibliography{example_paper}
\bibliographystyle{icml2026}

\newpage
\appendix
\onecolumn
\section{Appendix}
\subsection{Continuous Normalizing Flow Based VAE}
\label{appendix:cnfvae}

After projecting data into the latent space in a standard VAE architecture, the sampled latent variables are passed through the decoder to reconstruct the input. However, the representational capacity of this setup is inherently constrained by the Gaussian assumption. Even with a deep decoder, the model’s ability to capture complex, multi-modal, or highly non-linear latent structures is limited. To address this limitation, we introduce a Continuous Normalizing Flow (CNF) \cite{cnf} module that transforms the latent variables into a more expressive representation, effectively improving accuracy in the segmentation task. \textit{Normalizing Flows} \cite{nf} are a class of generative models that have received increasing attention due to their ability to construct flexible probability distributions through a sequence of invertible transformations with tractable likelihoods. CNFs extend this framework by defining the transformation as a continuous-time dynamical system governed by a parameterized ordinary differential equation: 
\begin{equation}
\frac{ \partial z^d(t)}{\partial t} = f(z^d(t), t), \quad z^d(0)=z^d_0,
\end{equation}
where $d\in\{s,t\}$ denotes the domain and $f$ is a neural network that defines a time-dependent vector field over the latent space. The network $f$ is assumed to be uniformly Lipschitz-continuous with respect to $z$, meaning that it is continuous transformation rather than a composition of discrete mappings. $z_0$ is the latent space variable produced by the encoder. 
This process maps the initial Gaussian distribution into a more complex and expressive distribution, from which the final latent representation $z^d_T = z^d(T)$ is sampled, where $[0, T]$ denotes the integration interval. Since the CNF also defines an invertible transformation of the latent variable, the resulting posterior density can be computed exactly using the change-of-variables formula, where the log-determinant of the Jacobian accounts for the change in volume induced by the flow:
\begin{equation}
    \frac{\partial \log p(z^d(t))}{\partial t} = - \text{Tr} \Big( \frac{\partial f}{\partial z^d(t)} \Big)
\end{equation}
Therefore, we can calculate the density of the distribution of the transformed variable $z^d_T$ as:
\begin{equation*}
    \label{eq:cnf_loglikeli}
    \log p(z^d_T) = \log p(z^d_0) - \int_{0}^{T} \text{Tr}\Bigg(\frac{\partial f}{\partial z^d(t)}\Bigg) dt
\end{equation*}
This formulation enables the transformation of the latent variable into a more expressive representation, while allowing for a more efficient calculation of its log-likelihood. 

\subsection{Gromov-Wasserstein Distance}
\label{appendix:gw}
In general, Optimal Transport Theory is based on the assumption that both distributions lie on the same metric space. However, there are many cases where direct comparisons between distributions are ill-defined or meaningless because the two distributions belong to different metric spaces, as is the case with comparing graphs or manifolds. In these cases, the Wasserstein distance cannot be defined due to the lack of a proper pointwise distance metric. The \textit{Gromov-Wasserstein} distance \cite{peyre2016gromov} addresses this limitation by comparing the internal structures of the metric spaces. The key idea is to find a coupling between the two probability distributions that minimizes the differences between the pairwise distances in each space. 

Formally, given two metric measure spaces $(X, d_X, \mu)$ and $(Y, d_Y, \nu)$, where $d_X, d_Y$ are distance metrics defined in each space and $\mu, \nu$ are probability measures, the Gromov-Wasserstein distance is defined as:

\begin{equation*}
    \mathrm{GW}^2(\mu, \nu) = \min_{\pi \in \Pi(\mu, \nu)} \int_{X^2 \times Y^2} \left| d_X(x, x') - d_Y(y, y') \right|^2 \, d\pi(x, y) \, d\pi(x', y')
\end{equation*}
This expression measures how well the two metric spaces can be matched under a coupling $\pi$ with marginals $\mu$ and $\nu$ by comparing how well the pairwise distances between their data points can be matched. By eliminating the need for a common metric space, the GW distance allows us more flexibility in areas like domain adaptation, where the source and target distributions may differ in dimensionality and feature space.

\subsection{Gaussian Gromov-Wasserstein Distance}
\label{subsec:HOT_GW}
Previous works often relied on the Euclidean distance to measure the cost of matching the latent representations of the source and target domains~\cite{JUMBOT, OLVA}. However, this assumes that the embedding space is isotropic, meaning that all dimensions contribute equally to the similarity of the embeddings, which is often unrealistic. To overcome this limitation, we adopt a more expressive distance formulation by considering each sample itself as a distribution, thanks to our VAE architecture that projects images into a Gaussian latent space. We leverage the Gaussian Gromov-Wasserstein (GGW) distance ~\cite{GGW}, which enables the comparison of embeddings with different latent space dimensions. This is particularly important in domain adaptation, because data from different domains inherently varies in complexity and structure, necessitating latent spaces of different dimensionality to effectively encode both domains. 

Specifically, the GGW distance is a closed-form version of the Gromov–Wasserstein distance when the distributions are Gaussian, such as the case with the VAE latent space embeddings. In \cite{GGW}, the authors show that there is no closed form solution for the Gromov-Wasserstein distance for Gaussian distributions, unless the optimal transport plan is also constrained to be Gaussian. By imposing a Gaussian structure on the transport plan, the problem can be reformulated in terms of matching the first- and second-order statistics of the distributions, yielding a tractable closed-form expression. In this case, given two Gaussian distributions $\mu \sim \mathcal{N}(\mu_s, \Sigma_s) \in \mathbb{R}^m$ and $\nu \sim \mathcal{N}(\mu_t, \Sigma_t) \in \mathbb{R}^n$ the set of acceptable transport plans becomes:
\begin{equation*}
    \gamma \in \Pi(\mu, \nu) \cap \mathcal{N}_{m+n}
\end{equation*}
which leads to the definition of the Gaussian Gromov-Wasserstein distance:
\begin{equation*}
\begin{aligned}
\mathrm{GGW}_2^2(\mu, \nu)
&=
4\Big(\operatorname{Tr}(D_s) - \operatorname{Tr}(D_t)\Big)^2 + 8\big\|D_s^{(n)} - D_t\big\|_F^2 + 8\Big(\|D_s\|_F^2 - \|D_s^{(n)}\|_F^2\Big).
\end{aligned}
\end{equation*}

where $D_s$ and $D_t$ are diagonal matrices containing the eigenvalues of $\Sigma_s$ and $\Sigma_t$, sorted in decreasing order and $\|\cdot\|_F$ denotes the Frobenius norm. We assume $m > n$, so the matrix $D_s^{(n)}$ corresponds to the submatrix formed by the leading $n$ eigenvalues of $D_s$, thus ensuring dimensional consistency.

\subsection{Unbalanced Gromov-Wasserstein Divergence}
\label{appendix:ugw}
The Gromov-Wasserstein distance is a formal way of comparing distributions on different metric measure spaces. However, in this formulation there is no room for creating or destroying mass from either distribution, which forces an exact solution that matches the respective marginals. This is problematic when the spaces have different sizes, contain outliers or noise etc. 

In order to generalize the Gromov-Wasserstein distance to these applications, the authors in \cite{séjourné2023unbalancedgromovwassersteindistance} introduce the \textit{Unbalanced Gromov-Wasserstein divergence}, which is a relaxation of the classic Gromov-Wasserstein formulation that allows the creation and destruction of mass. Formally, for two metric measure spaces $\mathcal{X} = (X, d_X,\mu)$ and $\mathcal{Y} = (Y, d_Y,\nu)$, it is defined as:
\begin{equation*}
    \text{UGW}(\mu, \nu) = \min_{\gamma \in \Pi(\mu, \nu)} \Big\{ \int_{X^2 \times Y^2} \left| d_X(x, x') - d_Y(y, y') \right|^2 \, d\gamma(x, y) \, d\gamma(x', y') +\epsilon_1 \text{D}_{\phi}^{\otimes} (\gamma_1|\mu) + \epsilon_2 \text{D}_{\phi}^{\otimes} (\gamma_2|\nu) \Big\}
\end{equation*}

where $\gamma_1, \gamma_2$ are the marginals of $\gamma$ in terms of the measures $\mu,\nu$ respectively, and $\epsilon_1, \epsilon_2$ the coefficients controlling the penalty terms for each marginal. The authors also introduce the quadratic $\phi-$divergence $\text{D}_{\phi}^{\otimes}$ between two probability measures $\rho, \kappa$, defined as:

\begin{equation*}
    \text{D}_{\phi}^{\otimes} (\rho||\kappa) \triangleq \text{D}_{\phi}(\rho \otimes \rho| \kappa \otimes \kappa) 
\end{equation*}

In our ablation studies, we include this formulation that acts as a replacement to the UOT formulation in Section~\ref{subsec:miniuot}. The point to compare the strengths and weaknesses of geometry-aware unbalanced transport against classical UOT, evaluating whether explicitly accounting for the metric structure of the data leads to measurable performance gains in domain adaptation.

\end{document}